\documentclass[runningheads]{llncs}

\usepackage{graphicx}
\usepackage[export]{adjustbox}
\usepackage{adjustbox}
\usepackage{array}
\usepackage{booktabs}
\usepackage{rotating}
\usepackage{multirow}
\usepackage{subcaption}
\usepackage{tikz}
\usepackage{comment}
\usepackage{amsmath,amssymb} %
\usepackage{color}
\usepackage{pifont}

\usepackage[accsupp]{axessibility}  %

\begin{document}
\pagestyle{headings}
\mainmatter
\def\ECCVSubNumber{100}  %

\title{Out-of-Distribution Detection \\ Without Class Labels} %

\titlerunning{Out-of-Distribution Detection Without Class Labels}
\author{Niv Cohen
\and
Ron Abutbul
\and
Yedid Hoshen
}
\authorrunning{N. Cohen, R. Abutbul, and Y. Hoshen.}
\institute{School of Computer Science and Engineering \\
The Hebrew University of Jerusalem, Israel \\
Correspondence to: \texttt{nivc@cse.huji.ac.il}
}
\maketitle

\begin{abstract}

Out-of-distribution detection seeks to identify novelties, samples that deviate from the norm. The task has been found to be quite challenging, particularly in the case where the normal  data distribution consists of multiple semantic classes (e.g., multiple object categories). To overcome this challenge, current approaches require manual labeling of the normal images provided during training. In this work, we tackle multi-class novelty detection \textit{without} class labels. Our simple but effective solution consists of two stages: we first discover ``pseudo-class'' labels using unsupervised clustering. Then using these pseudo-class labels, we are able to use standard \textit{supervised} out-of-distribution detection methods. We verify the performance of our method by a favorable comparison to the state-of-the-art, and provide extensive analysis and ablations.

\end{abstract}

\section{Introduction}

Detecting novelties, images that are semantically different from normal ones, is a key ability required by many intelligent systems. Some applications include: detecting unknown, interesting scientific phenomena (e.g. new star categories such as supernovae), or detecting safety-critical events (e.g. alerting an autonomous car when an unexpected object is encountered). In this paper, we assume that a training set entirely consisting of normal images is provided. A trained model is used at test time to classify new samples as normal or anomalous, i.e. similar or different from previously seen training samples. The normal data may consist of one or more semantic classes (e.g.``dog'', ``cat''), where the case of a single semantic class is of particular interest.

\begin{table}[h]
\begin{center}
\caption{Comparison of different novelty detection settings}
\label{tab:pam_comparison}
\begin{tabular}{cccc}
\toprule
 Normal Data  & One-class (OCC)	&	Multi-class (OOD)	 &	Our Setting 		\\
\midrule									
Multi-class 	& \ding{55}\footnote{1}	 &	\ding{51}	&	\ding{51}	\\
Without Labels 	& \ding{51}	 &	\ding{55}	&	\ding{51}	\\

\bottomrule
\end{tabular}
\end{center}
\end{table}

\footnotetext{A few novelty detection methods do evaluate on multi-class data without labels, and we compare to them in this work}

Most recent works that address out-of-distribution detection (OOD) %
have relied on \textit{supervised} class labels for each normal sample. On the other hand, methods that do not assume such labels have mostly focused on single-class data novelty detection. It is common to assume a normal-only dataset during training, so methods that do not assume class labels for this set are referred to as unlabelled methods. This setting of out-of-distribution detection with a normal only train set, but without class labels for the normal data, or \textit{multi-class anomaly detection} (AD), received little attention. . %

Here, we propose a simple approach for anomaly detection on unlabeled, multi-class normal only data. It is based on the following principle: unlabeled multi-class AD can be approximated using unsupervised clustering followed by supervised multi-class AD. In practice, we perform unsupervised image clustering using SCAN~\cite{van2020scan}, a recent state-of-the-art method. After obtaining the approximate labels for every image in the dataset, we can use them for adapting ImageNet-pre-trained features for the AD task. The adapted features are then used as input to generic AD methods e.g. $k$ nearest neighbor ($k$NN) or the Mahalanobis distance. 

Extensive experiments and ablations are performed to evaluate our method. We find that our method outperforms state-of-the-art self-supervised \cite{tack2020csi} and pre-trained \cite{bergman2020classification,reiss2021mean} anomaly detection methods. As an additional result, we show that the features learned by the unsupervised clustering method are already competitive with the best self-supervised anomaly detection methods.

Our main contributions are:

\begin{enumerate}

\item Presenting a framework that uses unsupervised clustering to obtain pseudo-labels which are then used to adapt feature representations for multi-class anomaly detection.

\item State-of-the-art results for unlabeled, multi-class anomaly detection on popular datasets.

\item Demonstrating that deep self-supervised image clustering methods are excellent feature learners for self-supervised multi-class anomaly detection.
\end{enumerate}

\subsection{Related Works}

\noindent\textbf{Out-of-Distribution Detection (OOD).} In some settings, the normal data are provided together with their semantic class ground truth labels. Using these labels for supervised training highly increases the model's ability to detect OOD samples \cite{hendrycks2016baseline}. A model trained to classify labelled semantic classes, may have less confidence in predicting the class label of a sample coming from a new distribution. Moreover, training a model to distinguish between semantic classes of the normal data enhances the sensitivity of the learnt features for the desired attribute. For example, a model trained to distinguish between animal species may learn a representation sensitive to their attributes (such as skin color and texture, head shape, etc.) and less sensitive to nuisance attributes (angle of view, lighting conditions, etc.). Using such representations, one can detect anomalous samples based on a large Mahalanobis distance in representation space, improving out-of-distribution detection capabilities \cite{lee2018simple,fort2021exploring}.

\noindent\textbf{One Class Classification.} In the absence of positive anomaly examples, anomaly detection methods design inductive biases allowing models to detect anomalies without seeing any during training. The common thread behind these methods is learning a strong representation of the data with which separating between normal and anomalous data is easily done.  %
Learning representations cannot be performed by using actual labels of the normal training data.  Therefore, anomaly detection algorithms, as other self-supervised method (e.g. image clustering \cite{van2020scan} or disentanglement \cite{kingma2013auto}), use a variety of techniques to design the inductive bias of the model toward having desired properties. Such techniques include self-supervised training to increase the model sensitivity to the properties essential for the task at hand \cite{hendrycks2019using,golan2018deep}, simulating samples which may resemble expected anomalies \cite{li2021cutpaste}, and data augmentation used to guide the model to ignore nuisance variation modes \cite{tack2020csi}. 

\noindent\textbf{Adaptation of Pre-trained Representations.} When pre-trained representations are available, they can be used to significantly improve anomaly detection accuracy. Early attempts   suggested a compactness loss to map the normal training data closer together, relying on auto-encoder pretraining \cite{ruff2018deep}. Similar techniques were adopted to ImageNet pre-trained features \cite{perera2019learning}, and later improved with self-supervised feature adaptation \cite{reiss2021panda,rippel2021transfer,reiss2021mean}. Pre-trained representations were lately adapted for the OE setting as well \cite{reiss2021panda,deecke2021transfer}.

\noindent\textbf{Clustering-based Anomaly Detection.} Most anomaly detection methods perform some form of density estimation of the normal data. When the normal training data is multi-class i.e. consists of multiple clusters, it is natural that the density model would include this inductive bias. Indeed, multiple clustering-based approaches have been used for anomaly detection including: K-means \cite{munz2007traffic} and GMMs \cite{li2016anomaly}. As clustering the raw data directly is unlikely to achieve strong results for images, deep learning methods have been applied to learn better representations, notably DAGMM \cite{zong2018deep}. In the past few years, clustering methods have not dominated anomaly detection benchmarks, even for multi-class data. In this paper, we revisit clustering-based anomaly detection and show that it achieves very strong performance for multi-class anomaly detection with and without pre-trained features. We note that SSD, a recently published work, uses K-means clustering for anomaly scoring in the setting of OOD without labels \cite{sehwag2021ssd}. We show that K-means clustering is less successful than our method on this task.

\noindent\textbf{Self-Supervised Image Clustering.} The task of clustering images without labels had been extensively studied. Most recent methods learn deep features simultaneously with optimizing the images cluster assignments \cite{caron2018deep}. Methods free to optimize their own features are prone to cluster according to nuisance properties, which the self-learnt feature might pick up.  A large variety of techniques have been used to mitigate this problem, including augmentations \cite{ji2019invariant} and contrastive feature learning \cite{van2020scan}.

\section{Preliminaries}
\label{sec:preliminaries}

In this section we present the necessary background for this paper.

\subsection{Deep Image Clustering}
\label{subsec:background:scan}

Recent years have seen significant progress in deep image clustering. This is mostly thanks to improvements in self-supervised representation learning. In this paper we take advantage of SCAN \cite{van2020scan}, a state-of-the-art unsupervised image clustering method. SCAN operates in three steps: 

\noindent \textit{(i) Representation learning.} SimCLR \cite{chen2020simple}, a contrastive self-supervised representation learning method, is used to learn a strong feature extractor $\phi_{SimCLR}$. 

\noindent \textit{(ii) Classifier training.} The feature extractor $\phi_{SimCLR}$ is used to compute the nearest neighbors for every training image. A classifier $C$ is trained (initialized with the features of $\phi_{SimCLR}$) to classify each training image $x$ into one of $K$ clusters. The classifier $C$ is trained under two constraints: (i) an equal number of images are assigned to each cluster. Formally, the entropy of $E_{x \in X_{train}}C(x)$ is maximized (ii) each image should be confidently assigned to a similar cluster as its nearest neighbours. Formally, we minimize $\sum_{\tilde{x} \in kNN(x)} log(C(x) \cdot C(\tilde{x}))$. By the end of training the classifier produces a clustering of the training set. 

\noindent \textit{(iii) Self-training.} In the final stage, the clustering is improved using a self-training approach \cite{mclachlan1975iterative}.

\subsection{Feature-Adaptation for Labeled Multi-Class Anomaly Detection}
\label{subsec:background:ood}

In the labeled multi-class anomaly detection task (also known as OOD), each normal training image $x$ is labeled by its class label $y$. Standard OOD methods train a classifier $C$ to predict the class probability vector $C(x)$ given $x$. The standard confidence-based score, measures for a test image $x_{test}$ the maximum prediction probability (probability of the most confident class):
\begin{equation}
    Confidence = \max\big({C(x_{test})}\big)
\end{equation}

Images with low maximal confidence, are denoted as anomalous. Many techniques have been proposed to improve the obtained OOD detection \cite{wei2022mitigating}. Here, we extend the work of Hendrycks et al. \cite{hendrycks2019using} and Fort et al. \cite{fort2021exploring} which suggested initializing the classifier $C$ with weights that are pre-trained on a large-scale datasets. %
This significantly increases the OOD detection accuracy. %
We propose a clustering-based technique for adapting these methods to unlabeled datasets.

\section{Deep Clustering for Multi-Class Anomaly Detection}
\label{sec:method}

\begin{figure}[t]
\centering
  \centering
    \begin{tabular}{c}
    \includegraphics[scale=0.28]{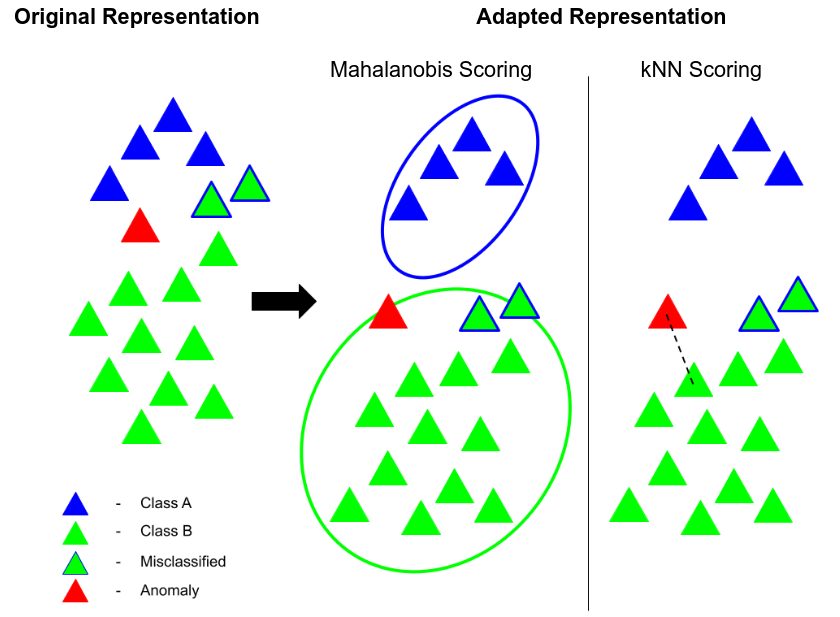}
    \end{tabular}
    \caption{An illustration of our feature adaptation and scoring procedure. Left - the data original unadpted pre-trained features. Feature adaptation increases the separation between the normal data (blue, green) and the anomaly (red). We illustrate that when labels are noisy w.r.t. the ground truth clusters, Mahalanobis distance scoring may misclassify anomalies (center, anomaly within the ellipses fitted to the normal data), while
    $k$NN scoring criterion (right, dashed line shows the anomaly's nearest in-distribution neighbour) may detect anomalies better.} 
    \label{fig:ood_clustering}

\end{figure}

In this work, we deal with anomaly detection when the normal training data is composed of many different semantic classes but no labels. We suggest taking advantage of the multi-class nature of the normal data as an inductive bias. The knowledge that the normal data comes from different, relatively distinct classes, rather than a single uni-modal class, or other possible distributions, serves us to learn better representations. As illustrated in Fig.\ref{fig:ood_clustering}, adapting features for better inter-cluster separation can provide better discrimination of anomalies, even when the cluster labels are imperfect. We first propose a naive but effective approach for using unsupervised clustering method for anomaly detection. We then propose a more accurate, two-stage approach which combines unsupervised clustering with pre-trained features and supervised multi-class anomaly detection for this task.

\subsection{Self-Supervised Clustering for Multi-Class Anomaly Detection}
\label{sec:method_self}

We first address the fully self-supervised case where no pre-trained features are available. We suggest a surprisingly simple but effective approach for this setting. First, we run the SCAN clustering algorithm (see Sec.\ref{sec:preliminaries}) on the normal training images. We than use  the features from the penultimate layer of the SCAN-trained classifier to represent each image. Anomalous samples are expected to be found in low-density area of this feature space. We estimate the normal samples density around a target image by its $k$NN distance to the normal training data.  We show in Sec.\ref{sec:res_self} that this simple method already achieves competitive results with the state-of-the-art methods when no auxiliary datasets are used. %

\subsection{Finetuning Pre-trained Features with Pseudo Labels}
\label{sec:method_pretrain}

\textbf{Pre-trained Features}. Using effective representations is at the core of deep anomaly detection techniques. It is well established by \cite{reiss2021panda,fort2021exploring}  that transferring deep representations learned on auxiliary, large-scale datasets (e.g. ImageNet or CLIP) is effective for anomaly detection. The main reason for their effectiveness is the ability to measure semantic similarity. In such representations, the normal data lies in relatively compact regions and are typically more separated from semantic anomalies.    The strong results from the fully self-supervise case above, motivate us to apply similar multi-class priors to pre-trained representation.

\noindent\textbf{Feature Adaptation.} Pre-trained representations are typically trained on large datasets, which are not necessarily representative of our normal in-distribution samples. Better AD results are obtained when the pre-trained features are adapted using the normal samples provided for training \cite{reiss2021panda}. A naive approach to adapt pre-trained features to our multi-class normal data is to finetune them using the loss of a clustering method, such as SCAN. %
However, we found that this form of feature adaptation results in a representation worse  than the initial one. We believe this is caused by catastrophic forgetting, where a network extensively trained for a new task loses capabilities and knowledge that it had during pretraining.

To overcome this limitation, we suggest a \textit{two stage approach}. %
In the \textit{first stage}, we simply train a self-supervised clustering method on the training data $X_{train}$ achieving approximate ``pseudo-labels'' describing clusters of our normal training data. In the \textit{second stage}, we transfer the knowledge obtained from the self-supervised clustering into the pre-trained network. To do this, we follow~\cite{fort2021exploring} and finetune our pre-trained network to classify training images into their pseudo-labels $\widetilde{y}_{train}$. %
We train the classifier  $C_{OOD}$, using standard cross entropy loss $\mathcal{L}_{CE}$: 

\begin{equation}
 \mathcal{L}_{CE} = - \sum_{i}  (\widetilde{y}_{train})_i \cdot \log \Big(C_{OOD}( X_{train} )_i \Big)
\end{equation}

\noindent\textbf{Model Averaging.} Although our two-stage approach is more stable than adaptation using the clustering objective directly, it can still lead to catastrophic forgetting. During finetuning, the network gains knowledge from the clusters we found, but forgets its pre-trained knowledge, which is crucial for the density estimation-based anomaly scoring we use. Although the model achieves its best performance on one particular epoch, it is non-trivial to select that epoch without relying on having anomalous samples during training. Therefore, we choose to score our anomalies with an average model, taken as the moving average of the weights of the model during different training epochs.

\noindent\textbf{Anomaly Scoring.}
To detect anomalies, we use density estimation with the adapted features.
Recent methods suggested a per-class Mahalanobis approach \cite{fort2021exploring} deeming a sample normal if it lies within a small Mahalanobis distance to any of the cluster centers. However, when the provided labels are not accurate this approach might fail. False cluster assignment may distort the Mahalanobis distance which relies on the empirical covariance matrix of each cluster. Instead, we use $k$NN scoring which makes few assumptions on the distribution of the data. Precisely, we encode all train images and the target images using the feature encoder. We then compute the distance between the features of each target image and all the normal train images. The anomaly score for each target image is set as the minimum of the distances to all nomral train samples. This score is relatively robust and less sensitive to the accuracy of the clustering algorithm (Fig.\ref{fig:ood_clustering}).

\section{Results}
\label{sec:results}

We evaluate our method on two commonly used OOD detection datasets. We use a variety of other datasets to simulate anomalies:

\textit{CIFAR-10}~\cite{krizhevsky2009learning}: Contains images from 10 different classes, with a 32x32 resolution. We evaluate it against a variety of dataset supplied in a similar resolution, namely: CIFAR-100~\cite{krizhevsky2009learning}, SVHN~\cite{netzer2011reading}, LSUN~\cite{yu2015lsun}. The hardest benchmark here is the CIFAR-100 which contain the most similar classes to CIFAR-10. For LSUN we report both a version with some artifacts used by previous works, and a version suggested by CSI \cite{tack2020csi}\footnote{https://github.com/alinlab/CSI} where the downscaling was done more carefully to avoid these artifacts. We do not compare on ImageNet\cite{deng2009imagenet} data here, as this dataset was used for pretraining.

\textit{ImageNet-30:} Contain images from 30 classes of high resolution images chosen from the ImageNet\cite{deng2009imagenet} dataset. Accordingly, we evaluate it against of a variety of datasets that have similar resolution, namely: CUB-200~\cite{WelinderEtal2010}, Dogs~\cite{deng2009imagenet}, Pets~\cite{parkhi12a}, Flowers~\cite{Nilsback08}, Food \cite{bossard14}, Places~\cite{zhou2014learning}, Caltech~\cite{fei2004learning}. %

\subsection{Multi-Class Anomaly Detection With Pre-trained Features}
\label{sec:res_pre-trained}

In this section we provide results for the case where pre-trained features are available. 

\textbf{Methods}: We evaluate anomaly detection methods that utilize pre-trained features, and do not require class labels for the train data. We present a comparison to a naive initialization of unsupervised clustering with pre-trained features in Sec.\ref{sec:discussion}.

\textit{Deep Nearest Neighbors (DN2)} \cite{reiss2021panda}: A $k$NN density estimation method based on pre-trained features. Test samples are scored according to the distance to their nearest normal training images. A larger distance indicates a low density of normal samples, and therefore a high probability of abnormality. We follow \textit{MeanShifted} \cite{reiss2021mean} by using the cosine similarity distance instead of $l_2$ distance.

\textit{Mean Shifted Contrastive Loss (MSCL)} \cite{reiss2021mean}: A-state-of-the-art method for adapting pre-trained features for anomaly detection using contrastive learning. It suggests to compute the contrastive loss around the mean of the normal training data features to mitigate catastrophic forgetting.

\textit{Ours:} The two stage clustering with pseudo-label adaptation method described in Sec.\ref{sec:method_self}. The hyperparameters and implementation details are described in Sec.\ref{sec:impl_det}.

\textbf{Comparison}: As can be seen in Tab.\ref{tab:ood_cifar10} and Tab.\ref{tab:ood_inet30}, feature adaptation has significantly improved our multi-class anomaly detection performance. A substantial part of our improvement comes from using large network architectures. We emphasize that the ability to easily use readily-available pre-trained networks is inherent to our approach.
Results were also improved in MSCL, but as MSCL is not explicitly designed to deal with multi-class data, our method significantly outperforms it. %
Using more powerful pretraining networks (as in \cite{fort2021exploring}) can also improve the results of our method. We find that ViT \cite{dosovitskiy2020image} pre-trained on ImageNet-21 \cite{kolesnikov2020big} achieves better results than ResNet152 pre-trained on ImageNet.
As ImageNet pretraining include the exact ImageNet class labels of the ImageNet-30 dataset we wish to avoid in our setting we used a CLIP \cite{radford2021learning} pre-trained ViT architecture for all pre-trained methods (including ours) in the ImageNet-30 evaluations.

\begin{figure}[t]
\centering
    \includegraphics[scale=0.27]{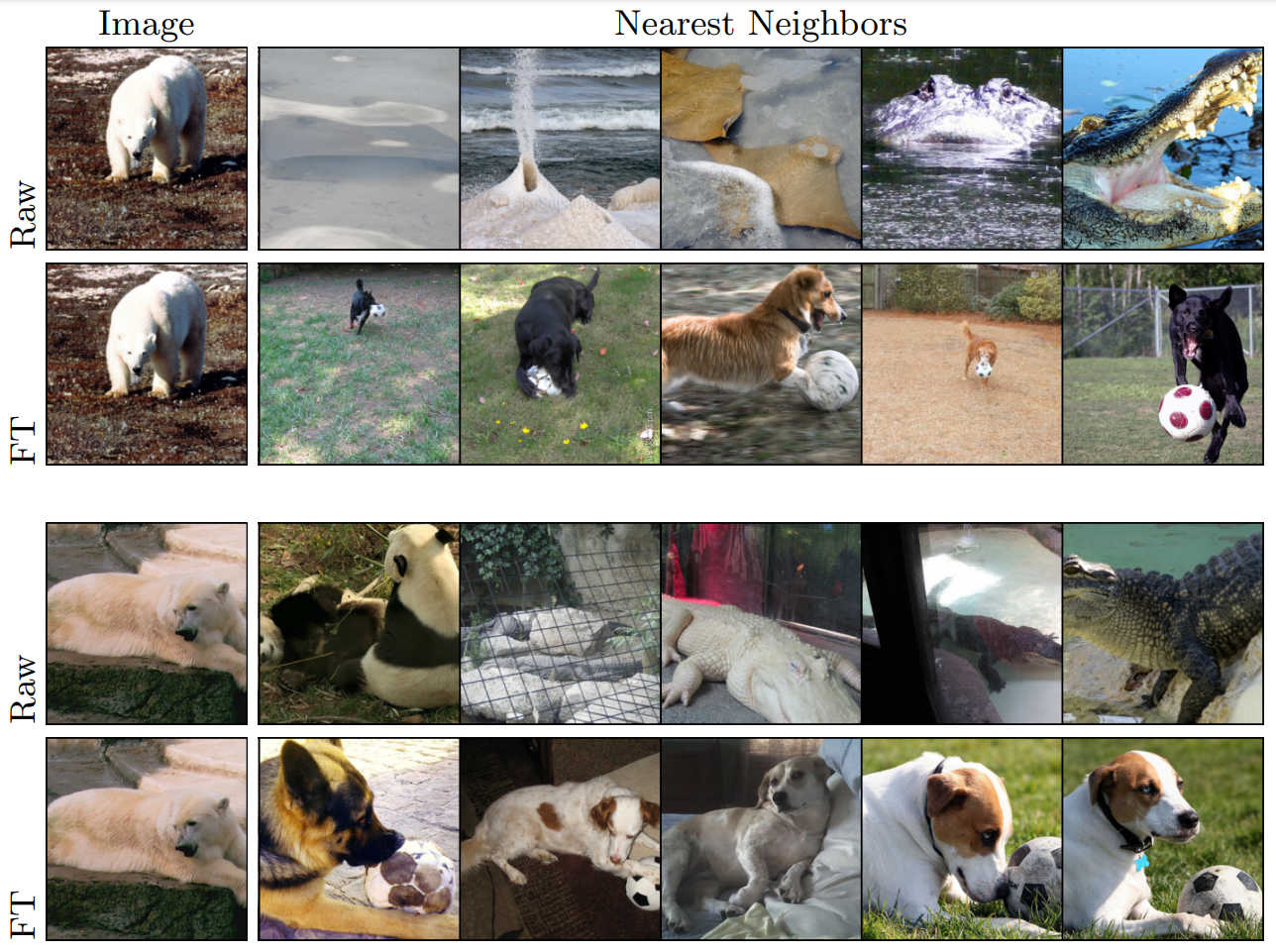}

\caption{\textit{\textbf{Raw pre-trained features vs. our finetuned features:}} For each out-of-distribution  image from the Caltech-256 dataset, the top $5$ nearest neighbors from the in-distribution data (ImageNet-30) are shown according to their order. Note how raw pre-trained features (Raw) neighbors are chosen from all classes. In contrast, finetuned pre-trained features with our method (FT) neighbors belong to a single semantic class.}
\label{fig:nn_images}
\end{figure}

When we compare between the $k$NN image retrievals of the original pre-trained features and our adapted features, we observe some differences. As can be seen in Fig.\ref{fig:nn_images} the raw pre-trained features retrieve images from different classes. These images are similar to our target out-of-distribution target sample in various semantic attribute. Adapted features, however, tend to associate our target image with a single normal class, resulting in a lower similarity between the target image and the retrieved normal ones. This is desirable, as we want to distinguish out-of-distribution images from our normal data.

\begin{table*}[t]
\caption{OOD detection without class labels on CIFAR-10 ROCAUC(\%) }
\label{tab:ood_cifar10}

\small

\begin{center}
\begin{tabular}{l  c c c c c c c c c c c c c c c c c c c}

\toprule
& & &	Network &	CIFAR-100 & SVHN	&	LSUN	&	LSUN  (FIX)		\\
\midrule																				
    \multirow{8}{*}{\rotatebox[origin=c]{90}{\scriptsize{\textbf{ }}}} & \multicolumn{2}{l}{Likelihood} 	& Glow	& 58.2	& 8.3	& -
&		- 
\\

& \multicolumn{2}{l}{Likelihood Ratio \cite{ren2019likelihood}} 	& PixelCNN++ & -	& 91.2	& -
&		- 
\\

& \multicolumn{2}{l}{Input Complexity \cite{serra2019input}} 	& Glow	& 73.6	& 95.0	& -
&		- 
\\

& \multicolumn{2}{l}{Rot \cite{hendrycks2019using}} 	& ResNet-18	& 82.3	& 97.8	& 92.8
&	81.6 
\\

& \multicolumn{2}{l}{GOAD \cite{bergman2020classification}} 	& ResNet-18	& 77.2	& 96.3	& 89.3
&	78.8 
\\

& \multicolumn{2}{l}{CSI \cite{tack2020csi}} 	& ResNet-18	& 89.2	& 99.8	& 97.5
&	90.3 
\\

 & \multicolumn{2}{l}{ SSD	\cite{sehwag2021ssd}} 	& ResNet-18	& 89.6	& -	& -
&	-
\\
	
 & \multicolumn{2}{l}{SCAN Features 	} 	& ResNet-18	& 90.2	& 94.3	& 92.4
&	92.1 
\\

\midrule		

& \multicolumn{2}{l}{	DN2 \cite{reiss2021panda}} & ResNet-18	& 83.3
&	95.0	&	91.0		&  88.9
\\

    \multirow{5}{*}{\rotatebox[origin=c]{90}{\scriptsize{\textbf{Pretrained}}}}  & \multicolumn{2}{l}{	DN2 \cite{reiss2021panda}}	& ResNet-152 	& 86.5
&	96.2	&	88.7 		& 86.7 
\\

& \multicolumn{2}{l}{	MSCL \cite{reiss2021mean}} & ResNet-152	& 90.0
&	98.6	&	90.6 		& 92.6 
\\
& \multicolumn{2}{l}{	Ours} & ResNet-18	& 90.8
&	98.6	&	98.6		& 94.3

\\
& \multicolumn{2}{l}{	Ours}	& ResNet-152	& 93.3
&	98.8	&	95.4 	& 95.7 
\\
& \multicolumn{2}{l}{	Ours}	& ViT	& \textbf{96.7}
&	\textbf{99.9}	&	\textbf{99.3} 		& \textbf{99.1 }
\\

\bottomrule

\end{tabular}
\end{center}
\end{table*}

\begin{table*}[t]
\caption{OOD detection without class labels on ImageNet-30 ROCAUC(\%).}
\label{tab:ood_inet30}

\small

\begin{center}
\begin{tabular}{l  c c c c c c c c c c c c c c c c c c c}

\toprule
& & &	Network &	CUB-200 & Dogs	&	Pets	&	Flowers-102 	&	Food-101 	&	Places-365 & Caltech-256  \\% & DTD 	\\ 
\midrule																				
    \multirow{3}{*}{\rotatebox[origin=c]{90}{\scriptsize{\textbf{ }}}}

& \multicolumn{2}{l}{Rot} & ResNet-18	& 74.5	& 77.8	& 70.0	& 86.3
&	71.6	&	53.1 & 70.0  \\% & 89.4 \\

& \multicolumn{2}{l}{GOAD} 	&  ResNet-18 & 71.5	& 74.3	& 65.5	& 82.8
&	68.7	&	51.0 & 67.4 & \\% 87.5 \\

& \multicolumn{2}{l}{CSI} & ResNet-18	& 90.5	& \textbf{97.1}	& 85.2	& 94.7
&	89.2	&	78.3 & 87.1 \\% & 96.9 \\

\midrule		

\multirow{2}{*}{\rotatebox[origin=c]{90}{\scriptsize{\textbf{Pretr.}}}}  & \multicolumn{2}{l}{	DN2}	& CLIP ViT	& 93.8
&	94.2	&	89.7 	& 93.2	& 95.7 & \textbf{96.7} & 90.3 \\% & \textbf{98.6} \\

& \multicolumn{2}{l}{	Ours}	& CLIP ViT	& \textbf{99.4}
&	95.9	&	\textbf{94.9} 	& \textbf{98.3}	& \textbf{96.4} & 96.1 & \textbf{94.4} & 
\\% \textbf{98.6} \\

\bottomrule

\end{tabular}
\end{center}
\end{table*}

\subsection{Multi-Class Anomaly Detection Without Pretraining}
\label{sec:res_self}

Although strong pre-trained features are often available, we cannot always assume their availability. In this case, we find that self-supervised clustering on its own can often learn a good enough representation to outperform previous state-of-the-art on unlabelled multi-class anomaly detection without pretraining.

\noindent\textbf{Methods}: We compare the following methods: 

\noindent\textit{Density based methods}: Classical methods use direct density estimation techniques to estimate the likelihood of the data. Different methods suggested modification to this score to account for the dataset statistics (\textit{Likelihood Ratio} \cite{ren2019likelihood}) or its complexity (\textit{Input Complexity} \cite{serra2019input}).

\noindent\textit{Rot} \cite{hendrycks2019using}: A classification based method utilizing an auxiliary task of rotation prediction for self-supervised detection of anomalies.

\noindent\textit{GOAD} \cite{bergman2020classification}: Another rotation-prediction method, that proposed to learn a feature space where inter-class separation of the normal data is relatively small.

\noindent\textit{CSI} \cite{tack2020csi}: A contrastive learning method which contrasts against distribution-shifted augmentations of the data samples along with other samples.

\noindent\textit{SSD \cite{sehwag2021ssd}:} A self-supervised method with similar features to CSI \cite{tack2020csi}. It scores anomalies using the Mahalanobis distance with respect to K-means clusters. Although this method is somewhat similar to ours, it relies on Mahalanobis distance and K-means clustering, which are not optimal for feature adaptation without labels (see Sec.\ref{sec:discussion}). We note that \textit{SSD\cite{sehwag2021ssd}} also reports higher results for the ResNet-34 architecture, which is a non-standard evaluation for AD without pretraining methods. 

\noindent\textit{SCAN Features} \cite{van2020scan}: Features taken from the last stage of of the SCAN clustering methods, used to score anomalies as in DN2 (see Sec.\ref{sec:method_pretrain}).

\noindent\textbf{Comparison}: As can be seen in Tab.\ref{tab:ood_cifar10}, simple utilization of SCAN features often performs better or on par with the top competing self-supervised method. We note that the last stage of the SCAN method, namely, self-labelling, is somewhat similar to adapting on pseudo-label performed by our method. Therefore, we do not expect our method to provide further gains over SCAN's final representation. %

Although clustering methods such as K-means and GMM have classically been very popular, most recent deep learning methods do not use clustering. The results reported here provide strong motivation for revisiting self-supervised clustering methods for anomaly detection. We conclude that relying on the multi-class distribution prior often allows us to outperform other methods, even in the setting where no pretraining is allowed.

\subsection{Implementation Details}
\label{sec:impl_det}

\noindent\textbf{Clustering:} We use SCAN's official implementation\footnote{https://github.com/wvangansbeke/Unsupervised-Classification} for all clustering tasks unless mentioned otherwise. We ran all of our clustering algorithms with the same number of clusters $K = 10$. %
We use the SCAN algorithm's default parameters for each dataset. For ImageNet-30 dataset we use the configuration originally provided by the authors for the ImageNet-50 dataset. We note that SCAN unsupervised image clustering use a MoCo\cite{chen2020improved} pretraining on the entire ImageNet dataset (pre-trained without labels). We therefore do not compare it to self-supervised methods that do not use pretraining \cite{hendrycks2019using,bergman2020classification,tack2020csi,sehwag2021ssd}.

\noindent\textbf{Pretraining:} For all models using ResNet152, we use ImageNet\cite{deng2009imagenet} pretraining. For ViT, we used ImageNet-21\cite{kolesnikov2020big} pretraining. To evaluate our method with ImageNet-30, we used the CLIP \cite{radford2021learning} ViT visual head rather than pretraining on labelled ImageNet data (Tab.\ref{tab:ood_inet30}).

\noindent\textbf{Optimization:} We ran the adaptation for $5$ epochs using an Adam optimizer. We used Cosine Annealing learning rate scheduler with an initial learning rate of $1e-5$ and final learning rate of $1e-6$.

\noindent\textbf{Model Averaging:} We average on the model weights at the end of each of the $5$ training epochs.

\noindent\textbf{Scoring}: We used $k = 1$ for all our $k$NN evaluations.

\noindent\textbf{Comparison to MSCL:} For the \textit{MSCL} comparison we experimented with $5,10$ and $100$ training epochs, and chose the best performing number of epochs. The rest of the parameters were left unchanged. %

\section{Discussion}
\label{sec:discussion}
\textbf{Do we actually need our two-stage approach?} We compare between pre-trained feature adaptation using two approaches: (i) Our two stage clustering with pseudo-class labels supervised adaptation (ii) A single-stage initialization of the unsupervised clustering method with pre-trained features. In the latter, we initialize the SimCLR features used in SCAN with ResNet152 ImageNet pre-trained features. The results are presented in Tab.\ref{tab:scan_stages_comparison}. The full faeture adaptation on the CIFAR-10 vs. CIFAR-100 evaluation results in $89.3$\% ROCAUC 
which is far worse than the $93.3$\% results achieved by our two-stage approach. During the clustering stage, the pre-trained features are used find the data clusters, performance but at the same time, the features deteriorate further away from their pre-trained initialization.
This justifies our two stage approach, which better preserves the expressivity of the pre-trained features.

\begin{table}[h]
\begin{center}
\caption{Comparison between OOD detection w.o. labels using the representations obtained by different stages of the SCAN clustering method (CIFAR-10 vs. CIFAR-100) ROCAUC (\%). The results show that the naive approach of directly using SCAN with pre-trained features falls behind our two-stage approach.	}
\begin{tabular}{cccc}
\toprule
	Ours  (ResNet-152) &	SIMCLR	&	SCAN & Self-Labelling		\\
\midrule									
	\textbf{93.3} & 87.3	&	83.4 	&	89.3	\\

\bottomrule
\end{tabular}
\label{tab:scan_stages_comparison}

\end{center}
\end{table}

\textbf{Is $k$NN density estimation preferable to other anomaly scoring methods?} Sometimes. Previous works used a variety of scoring criterion for multi-class anomaly detection \cite{ruff2018deep,rippel2021transfer}. Although the Mahalanobis distance was shown to give stronger results in a previous work, it is sensitive to the approximate labels that are typically generated by self-supervised clustering (see Fig.\ref{fig:ood_clustering} for an illustration). A comparison between the different scoring methods can be seen in Tab.\ref{tab:scoring_comparison}. We find that the optimal scoring method may differ between datasets. Following previous works \cite{cohen2020sub}, all our main results are reported using $k$NN with $k=1$.

\begin{table}[h]
\begin{center}
\caption{Comparison between scoring methods  ROCAUC (\%)	}
\label{tab:scoring_comparison}
\begin{tabular}{ccccccc}
\toprule
Scoring method	& 	$1$NN \quad	&	$2$NN	\quad &	$5$NN \quad	&	$10$NN	\quad &	 Mahalanobis & 	Confidence	\\
\midrule									
CIFAR10-CIFAR100 &	96.7 & 96.8 &	\textbf{96.9} & 96.8 & 96.6 & 92.9	\\
CIFAR10-SVHN & \textbf{99.9}	 & \textbf{99.9}	& \textbf{99.9} &	\textbf{99.9} &	92.4 &	97.1 \\
ImageNet30-CUB200 &	99.3 &	99.4 &	99.4 &	99.4 &	99.6 &	\textbf{99.7} \\
\bottomrule
\end{tabular}
\end{center}
\end{table}

\textbf{Can simple K-means clustering be used to finetune our features?} To a small extent. We evaluated the labels obtained by K means clustering as an alternative for the more complicated SCAN clustering method. Although the adaptation yielded small improvements, the results underperformed SCAN significantly. For example the multi-class anomaly detection results only improved from $86.5$\% ROCAUC to $87.3$\% (CIFAR-10 vs. CIFAR-100).

\begin{table}[h]
\begin{center}
\caption{Comparison of different numbers of clusters $K$ (ImageNet-30 vs. CUB200) ROCAUC (\%)}
\label{tab:cluster_no}
\begin{tabular}{ccccc}
\toprule
$K$	&	10	&	 20	&	30	&	No Adaptation \\	
\midrule									
	& \textbf{99.1}	&	98.3	&	98.9 & 93.8	\\

\bottomrule
\end{tabular}
\end{center}
\end{table}

\textbf{Can using a larger model with ImageNet pretraining assist the SCAN clustering performance?} Not significantly. We tried to initialize the SCAN algorithm with a ResNet152 pre-trained on ImageNet, but achieved only a minor improvement in the clustering accuracy. Although adaptation of pre-trained features using SCAN results in catastrophic forgetting, it is very likely that image clustering accuracy can be substantially improved with a method designed to take advantage of pre-trained representations. Such future improvements in image clustering are likely to directly enhance the results of our approach too.

\textbf{Does model averaging result in a good representation with comparison to individual checkpoints?} Yes. Model weight averaging can yield a similar (or better) multi-class anomaly detection accuracy compared to that of the optimal epoch along the training process. For example, on CIFAR-10 vs. CIFAR-100 with pre-trained ResNet152, the initial pre-trained features scored anomalies with $86.5$\% ROCAUC. The 5 individual training epochs scored $92.9\%$, $88.9\%$, $92.9\%$, $92.6\%$, $92.3\%$. The averaged model scored $93.3$\% ROCAUC. This phenomenon has also been noticed by previous works \cite{matena2021merging}. %

\textbf{What is the sensitivity of our method to random repetitions?}
We do not provide error bars for each of our runs since our results are fairly consistent. As a typical case, we ran 3 repetitions of our approach for the  CIFAR10-CIFAR100 and CIFAR10-LSUN (fix) experiments. The standard deviations were $0.4$\% and $0.5$\% respectively. 

\textbf{What is the sensitivity of our method to the number of clusters $K$?} While changing $K$ may somewhat vary the results, our method can often perform useful adaptation without knowing the exact ground truth number of classes (Tab.\ref{tab:cluster_no}). When available, we advise using the ground-truth number of clusters. Recently published methods have been able to infer the value of $K$ automatically \cite{ronen2022deepdpm}.

\textbf{How does adaptation with the pseudo-labels obtained by clustering compare to using the ground-truth labels?} There is still a significant gap between the pseudo-labels provided by our clustering algorithm and the ground truth class labels (on CIFAR-10, SCAN clustering accuracy is 88.3\%). Yet, the gap in the multi-class anomaly detection results seems to be smaller. For example, in the CIFAR-10 vs. CIFAR-100 multi-class anomaly detection task, our method achieves 96.7\% ROCAUC, while a similar method by Fort et al. \cite{fort2021exploring} achieves $98.4$\% ROCAUC using the ground truth labels. Even though this a significant difference, we find it is reasonable given the significant inaccuracy of self-supervised clustering we used.

\textbf{Relation to auxiliary-task based anomaly detection.} Our work can be seen as an extension of a line of methods utilizing auxiliary tasks to address a one-class-classification setting. Such methods were previously suggested for image anomaly detection \cite{golan2018deep}\cite{hendrycks2019using}, and also for other data modalities \cite{bergman2020classification}. These methods rely on predefined augmentations to create an auxiliary task in order to guide the model learning toward meaningful properties of our data. The prediction of pseudo-labels from clustering algorithms can be viewed as another example of such auxiliary work. Future works may suggest new kinds of data-adaptive auxiliary tasks for similar settings.

\textbf{New data modalities.} Multi-class anomaly detection may be encountered in other modalities beyond images. Data modalities where transfer learning and self-supervised learning show promising results include natural language, video and audio. Therefore, we believe similar methods may provide comparable improvements in these modalities. 

\textbf{Guided anomaly detection.} Another interesting application of clustering based out-of-distribution detection is using the pseudo labels to guide our anomaly detection algorithm toward the type of samples that we wish to consider as anomalies. For example, if one wishes to define anomalies according to colors, they may use augmentations to guide the clustering procedure toward finding color-based clusters. Similarly, one may wish to ignore an attribute, and guide the clustering algorithm accordingly.

\section{Limitations}
\label{sec:limitation}
\textbf{Highly unbalanced multi-class datasets.} Most state-of-the-art clustering algorithms rely on the assumption of an approximately equal split of the data among the classes. As our method relies on such algorithms, performance may decrease on imbalanced datasets. We expect that future clustering methods will overcome this limitations, consequently also freeing our method from this issue.

\textbf{Fine-grained and non-standard classes.} Many self-supervised learning algorithms, including self-supervised clustering, heavily rely on augmentations and other sources of inductive bias. The inductive bias in these methods guides the model to be sensitive to a single salient object in the center of the image. In cases when the anomalies or the semantic classes in the normal data are not object-centric, current self-supervised methods may not perform as well. While this is a limitation of our method, it is similarly a limitation of many other anomaly detection techniques reliant on self-supervised learning, including nearly all competing baseline methods. We therefore consider this as a limitation of the field in general, rather than a limitation specific to our method.

\textbf{Pre-trained features.} The main results of our paper are reliant on pre-trained features. As discussed, pre-trained features achieve strong results on most datasets, but may not be a good choice in some settings. One possible solution to this limitation, is combining pre-trained and fully self-supervised methods in scenarios where the preferred method cannot be determined in advance.

\section{Conclusion}
We address the problem of multi-class anomaly detection without labels. We propose a conceptually simple but effective method to combine recent improvements in unsupervised clustering with pre-trained features for anomaly detection. Our approach outperforms state-of-the-art feature adaptation methods for multi-class anomaly detection. Future work may utilize future improvements in self-supervised clustering of image datasets to further improve results. We also expect similar improvements in other data modalities such as natural language, video, audio and time-series.

\section{Acknowledgements}
This work was partly supported by the Malvina and Solomon Pollack scholarship and, the Federmann Cyber Security Research
Center in conjunction with the Israel National Cyber Directorate.

\clearpage
\bibliographystyle{splncs04}
\bibliography{egbib}
\end{document}